\documentclass[a4paper]{spie}

\usepackage{graphicx}
\graphicspath{{figures/}}
\usepackage{subfigure}

\usepackage{mathtools}
\usepackage{tabu}
\usepackage[colorlinks=true, allcolors=blue]{hyperref}

\title{The Parallel Algorithm for\\the 2-D Discrete Wavelet Transform}

\author{David Barina and Pavel Najman and Petr Kleparnik and Michal Kula and Pavel Zemcik}
\affil{Centre of Excellence IT4Innovations\\ Faculty of Information Technology\\ Brno University of Technology\\ Bozetechova 1/2, Brno\\ Czech Republic}

\begin{document}

\maketitle

\begin{abstract}
The discrete wavelet transform can be found at the heart of many image-processing algorithms.
Until now, the transform on general-purpose processors (CPUs) was mostly computed using a separable lifting scheme.
As the lifting scheme consists of a small number of operations, it is preferred for processing using single-core CPUs.
However, considering a parallel processing using multi-core processors, this scheme is inappropriate due to a large number of steps.
On such architectures, the number of steps corresponds to the number of points that represent the exchange of data.
Consequently, these points often form a performance bottleneck.
Our approach appropriately rearranges calculations inside the transform, and thereby reduces the number of steps.
In other words, we propose a new scheme that is friendly to parallel environments.
When evaluating on multi-core CPUs, we consistently overcome the original lifting scheme.
The evaluation was performed on 61-core Intel Xeon Phi and 8-core Intel Xeon processors.
\end{abstract}


\section{Introduction}

The two-dimensional discrete wavelet transform (DWT) is a very versatile image processing instrument.
It is employed in several image-compression standards (e.g., JPEG 2000).
As a consequence, many works deal with its fast implementation on all sorts of computer systems, including parallel architectures.
As it might be expected, many developers have adapted this transform on massively-parallel architectures, especially on GPUs.
However, all of these adaptations are based on the most popular separable schemes – the convolution and lifting schemes.
The separable convolution scheme can be computed in just two calculation steps, however, using a large number of arithmetic operations.
Whereas, the separable lifting scheme exhibits the smallest number of operations, and, on the contrary, the largest number of steps.
It is natural to expect that the number of operations should be proportional to a transform performance.
This is especially true on single-core CPUs.
However, it is essential that also the number of steps forms a bottleneck.
This is mainly meant in relation to multi-core processors.

In this paper, we show that the optimal scheme for multi-core CPUs lies aside the separable convolution and lifting schemes.
To the best of our knowledge, this problem has not been addressed in the literature yet.
The newly introduced scheme does not retain the separable property, as its operations cannot be associated with a horizontal or vertical direction.
In order to evaluate the proposed scheme, we performed several experiments on high-end server CPUs.
The evaluation is performed with CDF 5/3 wavelet, employed, e.g., in JPEG 2000 standard.
However, the presented schemes are general and they are not limited to any particular wavelet.

The rest of this paper is organized as follows.
Section 2 discusses a related work and introduces a mathematical notation used in the rest of the paper.
Section 3 presents the proposed non-separable scheme and its adaptation to a particular platform.
Section 4 evaluates the discussed schemes on multi-core CPUs.
Eventually, Section 5 summarizes and closes the paper.

\section{Background and Related Work}

This section introduces some notations and definitions to be used in the paper, and then it reviews conventional methods for computation of the \mbox{2-D} transform.

The widely-used $z$-transform is used for the description of wavelet filters.
Such filters are represented by polynomials in $z$ like $G(z)$.
Since this paper is focused on \mbox{2-D} transform, it is necessary to extend this notation into two dimensions.
So, two-dimensional filters look like $G(z_m, z_n)$, where the subscript $m$ refers to the horizontal axis and $n$ to the vertical axis.
The $G^*$ indicates a polynomial transposed to the original $G$.

The DWT splits the input signal into two components, according to a parity of its samples.
The components are often referred to as L and H.
The transform can be computed by a pair of quadrature mirror filters, referred to as G, followed by subsampling by a factor of 2.
Formally, this can be represented by the polyphase matrix
\begin{align}
	\label{eqn:1}
	\begin{bmatrix}
		G_1^{(o)} & G_1^{(e)} \\
		G_0^{(o)} & G_0^{(e)}
	\end{bmatrix}
	\text{,}
\end{align}
where operators $(e)$ and $(o)$ denote the even and odd terms of $G$.
This equation defines one-dimensional convolution scheme.
Further, Sweldens showed \cite{Daubechies1998} how the convolution scheme can be decomposed into a sequence of simple steps.
These filters are referred to as the lifting steps and the scheme as the lifting scheme.
The following paragraph discusses the lifting scheme in detail.

The initial polyphase matrix (\ref{eqn:1}) is factored into several pairs of lifting steps.
In each pair, the first step is called the predict step and the second one as the update step.
Formally, this can be represented by the product of polyphase matrices
\begin{align}
	\label{eqn:2}
	\prod_k
	\begin{bmatrix}
		1 & U^{(k)} \\
		0 & 1
	\end{bmatrix}
	\begin{bmatrix}
		1 & 0 \\
		P^{(k)} & 1
	\end{bmatrix}
	\text{,}
\end{align}
where $2K$ is the number of the lifting steps, and $P^{(k)}$ and $U^{(k)}$ represent the $k$th predict and update filter.
For simplicity, the superscript $(k)$ is omitted in the following text.

On multi-core CPUs, the processing of single or several adjacent signal samples is mapped to independent cores.
Due to the data exchange, the cores must use some synchronization method to avoid race conditions.
In the lifting scheme, these synchronizations can be required before the lifting steps.
In this paper, the synchronizations are indicated by the $|$ symbol placed before a polyphase matrix.
For example, $M_2 | M_1$ refers to a sequence of two lifting steps separated by some synchronization method.

\begin{figure}
	\hspace*{\fill}%
	\subfigure[]{\includegraphics{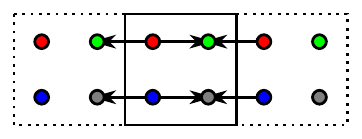}}%
	\hspace*{\fill}\\%
	\hspace*{\fill}
	\subfigure[]{\includegraphics{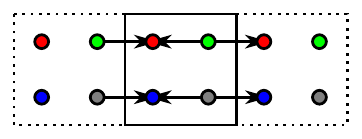}}%
	\hspace*{\fill}%
	\\[-128pt]%
	\hspace*{\fill}
	\subfigure[]{\includegraphics{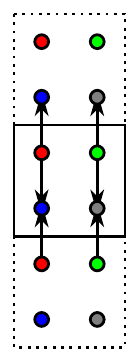}}%
	\hspace*{\fill}
	\subfigure[]{\includegraphics{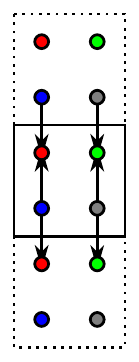}}%
	\hspace*{\fill}
	\caption{
		Shapes of steps in separable lifting scheme for the CDF 5/3 wavelet.
		The colored boxes correspond to LL, HL, LH, and HH quadruple.
		The arrows denote a multiply–accumulate operation.
	}
	\label{fig:1}
\end{figure}

Usually, the \mbox{2-D} transform \cite{Mallat1989} is defined as the tensor product of \mbox{1-D} transforms.
Unlike the \mbox{1-D} case, the \mbox{2-D} transform splits the input signal into a quadruple of wavelet coefficients (LL, HL, LH, and HH).
To describe \mbox{2-D} matrices, the predict and update operators must be extended into two dimensions.
Considering the separable lifting scheme, the predict and update lifting steps can be applied in both directions sequentially.
It should be noted that the horizontal and vertical steps can be arbitrarily interleaved.
The \mbox{2-D} lifting then follow from a sequence
\begin{align}
	\label{eqn:3}
	\left.\begin{bmatrix}
		1 & 0 & U^* & 0 \\
		0 & 1 & 0 & U^* \\
		0 & 0 & 1 & 0 \\
		0 & 0 & 0 & 1
	\end{bmatrix}\right|
	\left.\begin{bmatrix}
		1 & U & 0 & 0 \\
		0 & 1 & 0 & 0 \\
		0 & 0 & 1 & U \\
		0 & 0 & 0 & 1
	\end{bmatrix}\right|
	\left.\begin{bmatrix}
		1 & 0 & 0 & 0 \\
		0 & 1 & 0 & 0 \\
		P^* & 0 & 1 & 0 \\
		0 & P^* & 0 & 1
	\end{bmatrix}\right|
	\left.\begin{bmatrix}
		1 & 0 & 0 & 0 \\
		P & 1 & 0 & 0 \\
		0 & 0 & 1 & 0 \\
		0 & 0 & P & 1
	\end{bmatrix}\right|
	\text{.}
\end{align}
Note the synchronization $|$ before the matrices.
As the sequence can be hard to imagine, the individual matrices are illustrated in Figure \ref{fig:1} for the CDF 5/3 wavelet \cite{Cohen1992}.
For multiple lifting pairs, the scheme is separately applied to each such pair.
Recall that the separable lifting scheme has the smallest possible number of arithmetic operations and the highest number of steps.

Another scheme used for \mbox{2-D} transform is the separable convolution.
In this case, all calculations in a single direction are performed in a single step.
The drawback of this is the highest number of operations.
The scheme can formally be described as
\begin{align}
	\label{eqn:4}
	\left.\mathbf{N}^V\right|
	\left.\mathbf{N}^H\right|
	\text{,}
\end{align}
where $\mathbf{N}^H$ is a product of all steps in the horizontal direction and $\mathbf{N}^V$ is in the vertical one.
The convolution is followed by the subsampling.

So far, several studies have compared the performance of the separable lifting and convolution schemes on parallel architectures.
In an exemplary manner, the authors of \cite{Tenllado2008} compared these schemes on GPUs.
Although the results of their comparison are ambiguous, they concluded that the separable convolution is more efficient than the separable lifting counterpart in most cases.
They also claimed that fusing several consecutive steps might significantly speed up the execution, even if the complexity of the resulting fused step is higher.
In this regard, the authors failed to consider the possibility of a partial fusion, where the number of steps is reduced but it remains greater than a single step.
Other notable works can be found in \cite{Laan2011,Galiano2013,Song2014}.

This work is based on our previous work in \cite{Barina2016,Barina2017}.
In these papers, we introduced several non-separable schemes for calculation of \mbox{2-D} DWT suitable for graphics cards (GPUs).
We also presented a trick leading to a reduction of arithmetic operations.
The trick is also exploited in this paper.
In this paper, we extend previously presented schemes to multi-core CPU platform.
This is the point investigated in the following section.

\section{Proposed Scheme}

This section presents non-separable schemes suitable for multi-core CPUs.
A contribution of the paper starts with this section.

\begin{figure}
	\hspace*{\fill}%
	\subfigure[]{\includegraphics{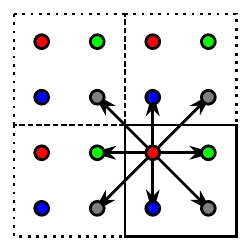}}%
	\hspace*{-40pt}%
	\raisebox{-32pt}{\subfigure[]{\includegraphics{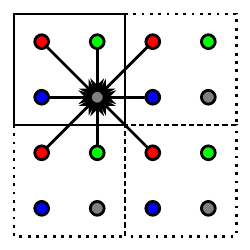}}}%
	\hspace*{\fill}%
	\subfigure[]{\includegraphics{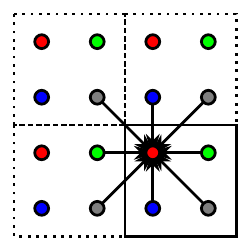}}%
	\hspace*{-40pt}%
	\raisebox{-32pt}{\subfigure[]{\includegraphics{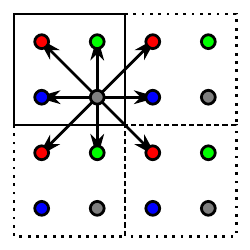}}}%
	\hspace*{\fill}%
	\null%
	\caption{
		Non-separable lifting scheme for the CDF 5/3 wavelet.
		The predict step on the left, update on the right.
	}
	\label{fig:2}
\end{figure}

The above-described approaches did not exploit the possibility of a fusion of polyphase matrices.
Having this in mind, all horizontal and vertical calculations of the corresponding pair of matrices can be performed in a single step.
The drawback of this approach is a higher number of operations and memory accesses.
Since CPUs are very sensitive to the total number of arithmetics operations, the fusion will be appropriate to apply to the lifting scheme.
In this way, non-separable lifting scheme is formed.
The scheme has the same number of steps as its separable counterpart.
On the other hand, the number of operations has been increased.
The scheme consists of a spatial predict and spatial update steps.
For curiosity, The predict step is completely responsible for the HH coefficient, whereas the update step for the LL one.
Formally, the scheme is defined by
\begin{align}
	\label{eqn:5}
	\left.\begin{bmatrix}
		1 & U & U^* & UU^* \\
		0 & 1 & 0 & U^* \\
		0 & 0 & 1 & U \\
		0 & 0 & 0 & 1
	\end{bmatrix}\right|
	\left.\begin{bmatrix}
		1 & 0 & 0 & 0 \\
		P & 1 & 0 & 0 \\
		P^* & 0 & 1 & 0 \\
		PP^* & P^* & P & 1
	\end{bmatrix}\right|
	\text{.}
\end{align}
The $PP^*$ and $UU^*$ are the spatial filters (tensor products of \mbox{1-D} filters).
For CDF 5/3 wavelet, it is illustrated in Figure \ref{fig:2}.

\begin{figure}
	\medskip
	\hspace*{\fill}%
	\subfigure[]{\includegraphics{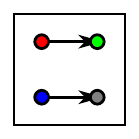}}%
	\hspace*{\fill}%
	\subfigure[]{\includegraphics{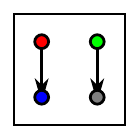}}%
	\hspace*{\fill}%
	\subfigure[]{\includegraphics{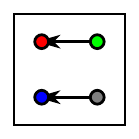}}%
	\hspace*{\fill}%
	\subfigure[]{\includegraphics{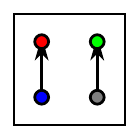}}%
	\hspace*{\fill}%
	\null\\%
	\hspace*{\fill}%
	\subfigure[]{\includegraphics{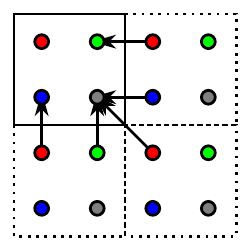}}%
	\hspace*{\fill}%
	\subfigure[]{\includegraphics{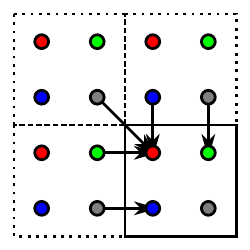}}%
	\hspace*{\fill}%
	\null%
	\caption{
		The platform-adapted non-separable scheme for the CDF 5/3 wavelet.
		The predict on the left, the update on the right.
	}
	\label{fig:3}
\end{figure}

As mentioned above, an optimization approach can adapt the schemes to a particular platform.
The number of operations or memory accesses can be reduced, while the number of computational steps remains unaffected.
Regardless the underlying platform, an important observation can be made.
A special form of the operations guarantees that the CPU cores never access the results belonging to extraneous cores.
These operations comprise constants (monomials with the zero exponents).
As the convolution is the linear operation, these polynomials can be detached from the original operations, and calculated using the separable scheme (due to the lowest number of operations).
Such schemes are referred to as adapted schemes.
Formally, the original polynomials are split as $P = P_0 + P_1$ and $U = U_0 + U_1$, where $P_0$ and $U_0$ are the desired constants.
The $P_1$ and $U_1$ are kept in the original non-separable scheme.
For a better understanding, the adapted non-separable scheme is illustrated in Figure \ref{fig:3}.

\section{Evaluation}

Since the above-listed properties do not provide sufficient information on performance in real environments, the performance on real multi-core CPUs is compared in this section.

In order to evaluate the considered schemes, high-performance server CPUs were used, along with the code written using the C language and OpenMP interface.
The evaluation was performed primarily on Intel Xeon and Intel Xeon Phi server processors.
Their technical parameters are summarized in Table \ref{tab:1}.
In the following paragraphs, several experiments on these CPUs are presented.

\begin{table}
	\caption{Description of the CPUs used for the evaluation.}
	\label{tab:1}
	\centering
	\medskip
	\begin{tabu} to \linewidth {X[l]X[l]X[l]}
		& Intel Xeon & Intel Xeon Phi (MIC) \\
		\hline
		model & Xeon E5-2620 v4 & Intel Xeon Phi 7120P \\
		cores & 8 & 61 \\
		concurrent threads & 16 & 244 \\
		clock (turbo) & 2.1 GHz (3.0 GHz) & 1.238 GHz (1.333 GHz) \\
		on-chip memory & 20 MB (L3 cache) & 30.5 MB (L2 cache) \\
		off-chip memory & DDR4 (2.133 GHz) & GDDR5 (2.75 GHz) \\
		\hline
	\end{tabu}
\end{table}

\begin{figure}
	\medskip
	\includegraphics[width=.5\linewidth]{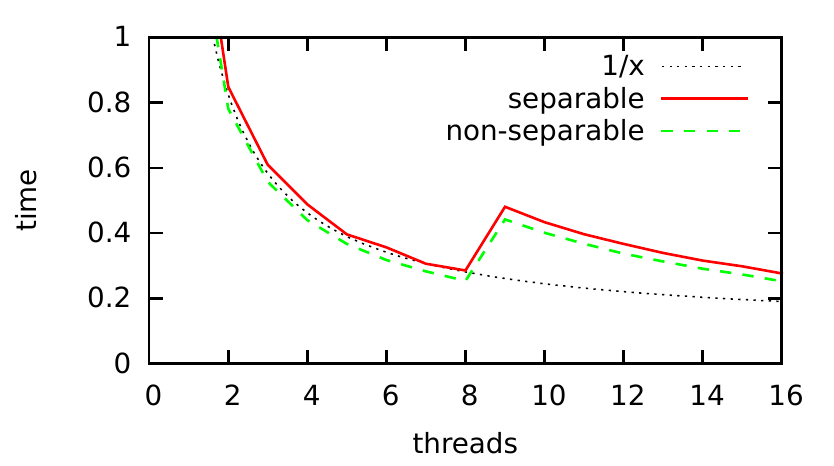}%
	\includegraphics[width=.5\linewidth]{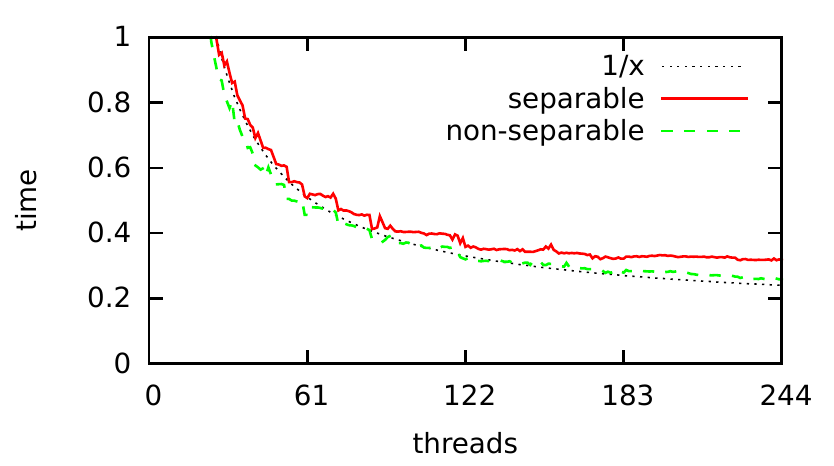}
	\caption{
		Examination of the optimal number of threads.
		The Xeon on the left, Xeon Phi on the right.
		The performance scales almost linearly.
	}
	\label{fig:4}
\end{figure}

In the first experiment shown in Figure \ref{fig:4}, the optimal number of threads was examined.
The measurements were conducted with separable and non-separable schemes and CDF 5/3 wavelet.
The transform performance was measured with tiles of $1024\times{}1024$ size, comprising single-precision floating-point values.
The presented results are a median of 100 measurements.
The time is given in nanoseconds per pixel (ns/pel).
It is clear from the figure that the curves roughly approximate the $1/x$ function, where $x$ is the number of threads.
Therefore, the measurements made show that the optimal number of threads roughly corresponds to the maximum number of threads available.
Note the phenomenon that occurs when the number of threads exceeds the number of CPU cores (i.e.
8 for the Xeon, 61 for the Xeon Phi).

\begin{figure}
	\includegraphics[width=.5\linewidth]{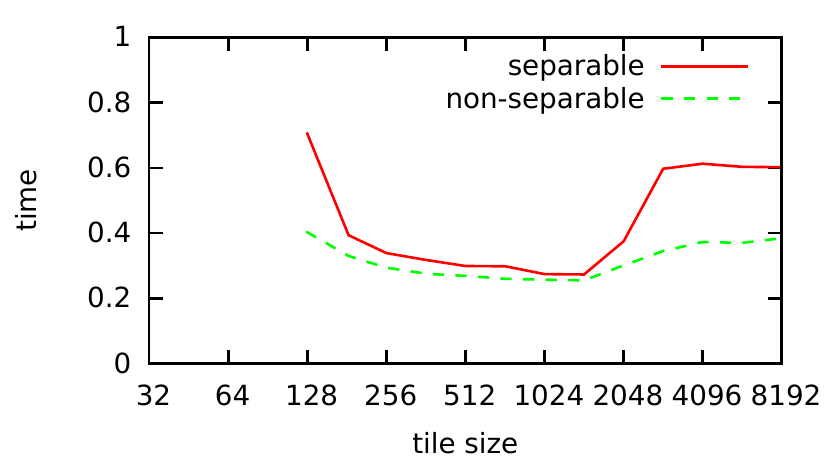}%
	\includegraphics[width=.5\linewidth]{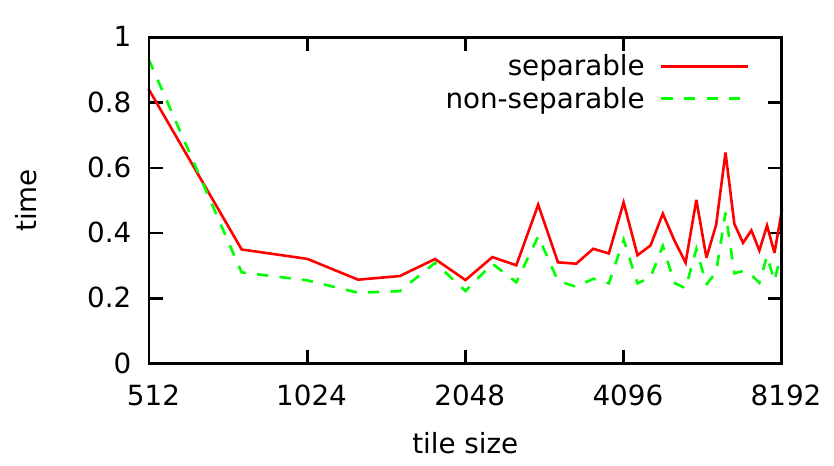}
	\caption{
		Examination of the optimal transform tile size.
		The Xeon on the left, Xeon Phi on the right.
	}
	\label{fig:5}
\end{figure}

In the second experiment in Figure \ref{fig:5}, the optimal transform tile size was examined.
The number of threads that was found optimal in the previous experiment was used.
For the Xeon CPU, the optimal power-of-two tile size was chosen as $1024\times{}1024$.
For the Xeon Phi, the size $2048\times{}2048$ was chosen.
Note that the tile size does not necessarily have to be a power of two, but this is a suitable choice, for example, due to JPEG 2000.

\begin{figure}
	\includegraphics[width=.5\linewidth]{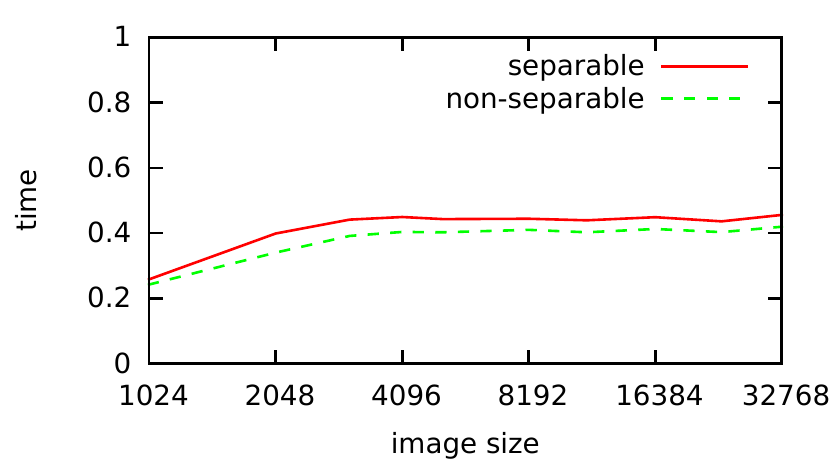}%
	\includegraphics[width=.5\linewidth]{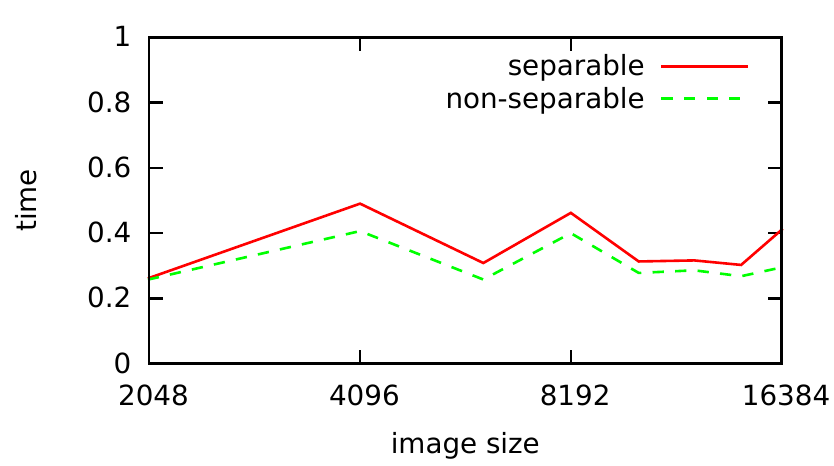}
	\caption{
		Performance on large images.
		The Xeon on the left, Xeon Phi on the right.
	}
	\label{fig:6}
\end{figure}

In the last experiment in Figure \ref{fig:6}, we were interested in a real performance.
The $x$-axis shows the size of the image edge.
The input and output images were supplied by the main memory.
Note that the image sizes exceed CPU cache size after a while.
The experiment confirms that the non-separable scheme consistently overcomes the original separable lifting scheme.
For example, for $8192\times{}8192$ image, the speedup factor is about 10\% on the Xeon and 25\% on the Xeon Phi processor.

In summary, we can conclude that the reduction in transform steps can improve performance, at least on some platforms.
All the source codes used in this article together with all the results are available in a repository \cite{supplementary-materials} on the website of the authors' affiliation.

\section{Summary}

This paper introduces and discusses the non-separable lifting scheme for computation of the two-dimensional discrete wavelet transform on multi-core CPUs.
We found that the non-separable scheme outperforms its separable counterpart in most cases.
We can confirm that fusing consecutive steps of the original lifting scheme might speed up the execution, irrespective of its higher complexity in terms of arithmetics operations.
The presented scheme is general and it can be used in conjunction with any wavelet transform.

For future work, we plan to extend our approach to other wavelets and possibly other non-separable schemes.
The implementation can also further be improved using appropriate SIMD extensions.
Finally, we look for other multi-core platforms such as multi-core ARM processors.

\acknowledgments

This work has been supported by
the Ministry of Education, Youth and Sports of the Czech Republic from the National Programme of Sustainability (NPU II) project IT4Innovations excellence in science (LQ1602),
and the Technology Agency of the Czech Republic (TA CR) Competence Centres project V3C – Visual Computing Competence Center (no. TE01020415).

\enlargethispage{\baselineskip}
\bibliographystyle{spiebib}
\bibliography{sources}

\end{document}